\documentclass{article}

\usepackage{iclr2027_conference,times}
\usepackage{amsmath,amssymb,booktabs,graphicx,multirow,siunitx}
\usepackage[table]{xcolor}
\usepackage{hyperref}
\usepackage{url}
\usepackage{xspace}
\usepackage{placeins}
\usepackage{fancyhdr}

\graphicspath{{evidence/figures/}}
\fancypagestyle{plain}{
    \fancyhf{}
    \renewcommand{\headrulewidth}{0pt}
}
\fancypagestyle{empty}{
    \fancyhf{}
    \renewcommand{\headrulewidth}{0pt}
}

\newcommand{\mv}[2]{%
#1\,\smash{\raisebox{-0.45ex}{\textcolor{gray!70}{\scriptsize #2}}}%
}
\definecolor{oursblue}{RGB}{220,236,246}

\newcommand{\method}{PreActIR\xspace}

\newcommand{\tablecaptiongap}{\vspace{4pt}}

\title{\raggedright
Look Before You Restore: Learning to Predict
Before Acting for Agentic Image Restoration
}

\author{
Xu Zhang$^{1}$ \quad
Jun Wan$^{2}$ \quad
Lefei Zhang$^{1,*}$ \quad
Liangpei Zhang$^{1}$ \\[6pt]
$^{1}$Wuhan University \\
$^{2}$Zhongnan University of Economics and Law \\[6pt]
\texttt{zhangx0802@whu.edu.cn}
}

\iclrfinalcopy

\begin{document}

\maketitle

\begingroup
\renewcommand{\thefootnote}{\fnsymbol{footnote}}
\footnotetext[1]{Corresponding author.}
\endgroup

\begin{abstract}
Agents for mixed-degradation image restoration typically learn what a tool does to an image only after running it: failed calls waste computation and harmful outputs go unnoticed until too late. We propose \textbf{PreActIR}, an agentic restoration framework that predicts tool consequences before execution. For each candidate tool, a tool-conditioned world model forecasts changes in residual degradation, image quality, and fidelity, together with the risk of harm and the uncertainty of the prediction. The agent executes the top-ranked candidate, and an independent verifier commits or rolls back its output before the agent replans from the observed state. PreActIR also audits whether performance is limited by the selector or by the candidate space, and studies targeted tool adaptation as a controlled way to expand that space.
In a controlled evaluation that fixes the candidate pool, verifier, and budget and employs reference-annotated degradation states, PreActIR uses 42.5\% fewer real tool calls and 63.4\% fewer rejected calls than reactive, execute-then-evaluate trial and error, while improving Y-PSNR by 0.92\,dB.
\end{abstract}

\section{Introduction}  

Image restoration has advanced rapidly for individual degradations
such as noise, blur, rain, haze, low light, compression, and low resolution~\citep{zhang2017dncnn,zhang2018ffdnet,wei2018retinexnet,chen2018sid,zamir2020mirnet,chen2021hinet,zamir2021mprnet,zamir2022restormer,chen2022nafnet,liang2021swinir,tu2022maxim,wang2021realesrgan,zhang2021bsrgan}. In practice, however, several degradations often coexist and
interact, making restoration a sequential decision problem in which each
intervention changes both the image state and the value of later actions.
Recent agentic systems coordinate specialized tools through
perception, planning, execution, and feedback~\citep{agenticir2025,zuo20254kagent,hybridagent2026,skillir2026}.
Most of them follow an execute--observe--replan loop~\citep{yao2023react}:
a tool's effect on a given image is known only after it has run, failed calls
still cost computation, and harmful side effects go unnoticed until
afterwards.

\begin{figure}[t]
    \centering
    \includegraphics[width=\linewidth]{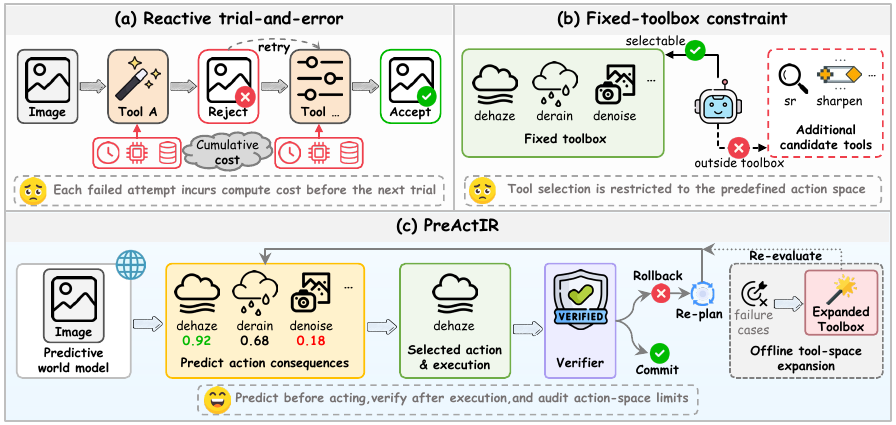}
\vspace{-2em}
    \caption{Motivation of \method. Existing agentic restoration is largely
    reactive: tool effects are observed only after execution, while decisions
    remain restricted to the available action space. \method predicts action
    consequences before execution.}
    \label{fig:motivation}
\vspace{-1em}
\end{figure}

{Recent agents reduce the number of trials with trained planners and stored
experience}~\citep{jia2026tiragent,zhu2026opera,wang2026paagent},
and a few methods score actions in advance: DiTTo~\citep{ditto2026} learns a
simulator and a per-action quality predictor for trajectory construction, and
the concurrent Causal-AgentIR~\citep{gao2026causalagentir} estimates plan
utility from a causal memory of past executions. Neither trains a
tool-conditioned consequence model for deployment-time decisions. {A separate
question is whether the tool set itself is the limit}: even a perfect selector fails if no reachable
tool sequence yields a good result. Agents can add experts~\citep{ditto2026}
or fine-tune tools together with the planner~\citep{zhu2026opera}, but a poor
result alone does not tell whether the selector chose badly or no good
candidate was reachable---two cases that call for different fixes. As
illustrated in Fig.~\ref{fig:motivation}, {reactive execution pays for an
outcome before seeing it, and it is usually unclear whether the selector or the
tool set is the binding limit.}

We propose \textbf{PreActIR}, an agentic restoration framework built around
the idea that an agent should predict what a tool will do before calling it.
A tool-conditioned world model takes each candidate tool and predicts the
changes in residual degradation, image quality, and fidelity, along with
content damage, the probabilities of benefit and harm, and the uncertainty of
these predictions. The agent ranks candidates by these predictions and
executes only the top one. Because predictions can be wrong, an independent
post-execution verifier checks the actual output and commits or rolls it
back; the agent then replans from the observed state. We call this the
predict--act--verify loop.

A separate diagnostic {asks whether the selector or the tool set is binding}. An offline
candidate-frontier audit uses test references post hoc to find the best result
reachable from the fixed candidate pool, separating selector headroom from
tool-space limits. We then adapt one existing tool on external replay data and
measure how it shifts both the attainable frontier and the deployed system.

Our contributions are as follows:
\begin{itemize}

    \item A \textbf{tool-conditioned world model} that predicts how each
    candidate tool changes residual degradation, quality, and fidelity, along
    with {benefit and harm probabilities and uncertainty, so
    that tools can be ranked and their risk estimated before execution.}

    \item A \textbf{predict--act--verify} loop that executes only the
    top-ranked tool {and lets} an independent verifier commit or roll back its
    output; in a matched replay {with reference-annotated states, it makes}
    42.5\% fewer real calls and 63.4\% fewer {rejections} than reactive
    trial and error.

    \item A \textbf{candidate-frontier audit} that distinguishes tool-selection limitations from candidate-space constraints by characterizing attainable fidelity–perception trade-offs. It quantifies the headroom available through better selection within a fixed pool and examines how targeted tool adaptation expands the attainable frontier.

\end{itemize}

\section{Related Work}
\label{sec:related}


\paragraph{All-in-One and Agentic Image Restoration.}
All-in-one image restoration addresses multiple degradations within a unified
model using degradation-aware representations, multimodal guidance, or adaptive
expert routing~\citep{airnet2022,promptir2023,mioir2024,daclip2024,
instructir2024,perceiveir2025,autodir2024,adair2025,moceir2025,clearair2026}.
These methods learn a fixed policy without predicting tool effects before
execution. Agentic restoration instead coordinates specialized models through
perception, planning, execution, and feedback, building on
reinforcement-learned toolchains~\citep{yu2018rlrestore,yu2021pathrestore}.
MLLM controllers reason about degradations and coordinate
experts~\citep{restoreagent2024,jarvisir2025,zhou2025qagent,wei2026iamagent};
AgenticIR, MAIR, 4KAgent, SEAR, HybridAgent, and FAPE-IR add exploration
with rollback, multi-agent or recursive execution, tree search, fast--slow
coordination, and frequency-aware planning~\citep{agenticir2025,mair2026,zuo20254kagent,sear2026,hybridagent2026,liu2026fapeir};
and others target rain, inscriptions, or photo
retouching~\citep{yu2026derainagent,zhu2026epiagent,xiong2026beyond}.
In these systems a tool's effect is observed only after it runs, and SEAR's
tree search scores candidates by real rollouts.

\paragraph{Learned Tool Selection and Pre-Execution Estimation.}
Restore-R1, TIR-Agent, and OPERA train the planner with perceptual or metric
rewards~\citep{lu2026restorer1,jia2026tiragent,zhu2026opera}, and OPERA also
fine-tunes the tools; SkillIR reuses skills from past
executions~\citep{skillir2026}, and PaAgent retrieves tool-specific experience
from a portrait bank of restoration tools~\citep{wang2026paagent}. These methods improve the choice of tool but do not predict what it will do. DiTTo~\citep{ditto2026} learns a
restoration simulator and a per-action quality predictor, but uses them to
build order-aware training trajectories. The concurrent
Causal-AgentIR~\citep{gao2026causalagentir} keeps a causal memory graph of past
executions, curated by a learned memory-operation policy, and scores plans
before execution by multi-hop reasoning over retrieved memory paths, with a
critic and rollback afterwards. Pre-execution scoring and rollback are therefore
not new on their own. \method differs in that it trains a consequence
model on real transitions and queries it for every candidate at
deployment time, predicting separate outcomes rather than collapsing them into
a single score; it also audits whether the selector or the tool set limits performance.

\paragraph{World Models and Predict-before-Act Agents.}
World models learn action-conditioned dynamics so that an agent can evaluate
actions in imagination before acting~\citep{ha2018worldmodels,hafner2019planet}.
Probabilistic models quantify their own uncertainty~\citep{chua2018pets},
short rollouts limit compounding model error~\citep{janner2019mbpo}, and a
model need not reconstruct observations if it predicts only decision-relevant
quantities~\citep{schrittwieser2020muzero}. Among language agents, PreAct
predicts action results before acting~\citep{fu2025preact}, and WebDreamer
uses an LLM to simulate each candidate web action before choosing
one~\citep{gu2025webdreamer}. Those predictions are free-form text from a
pretrained model, whereas restoration tool outcomes are continuous and
multi-dimensional, and fidelity to the unknown clean image is unobservable at
test time. Our consequence heads are instead supervised with uncertainty
estimates on real tool transitions; we plan one step ahead by default and rely on a
post-execution verifier to catch the remaining prediction errors.

\section{Method}

\subsection{Problem Formulation}

Given a degraded image $I_0$, a restoration-tool set $\mathcal A$, and a
real-execution budget $B$, we formulate agentic image restoration as a
budgeted sequential decision problem. At step $t$, the agent maintains a
restoration state
\begin{equation}
s_t = \big[f(I_t),\, \mathbf z_t^{\rm deg}\big],
\end{equation}
where $f(I_t)$ denotes the visual representation of the current image and
$\mathbf z_t^{\rm deg}$ characterizes its residual degradations. A candidate
generator constructs a feasible action set
\begin{equation}
\mathcal A_t = G(\mathbf z_t^{\rm deg},\mathbf M_t;\mathcal A),
\end{equation}
where $\mathbf M_t$ are spatial degradation maps (Sec.~\ref{sec:world-model});
the agent selects $a_t\in\mathcal A_t$ or terminates.

Executing $a_t$ invokes the corresponding restoration operator $T_{a_t}$ and
consumes one unit of the real-execution budget, even if the output is rejected
and rolled back ($I_{t+1}=I_t$). Let $N_T$ denote the total number of real tool
calls and $C(a)$ a relative execution-cost proxy.
The restoration objective is
\begin{equation}
\max_{a_{0:T-1}}\;
\Phi\!\left(\mathbf q(I_T)\right)
-\lambda_c\sum_{t=0}^{T-1} C(a_t),
\qquad
\text{s.t. } N_T \le B,
\label{eq:budgeted-objective}
\end{equation}
where $\Phi$ maps the quality measures $\mathbf q(I_T)$ of the final image to a
terminal utility. The difficulty is that an action's outcome is unknown until
its tool has run.

\begin{figure}[t]
    \centering
    \includegraphics[width=\linewidth]{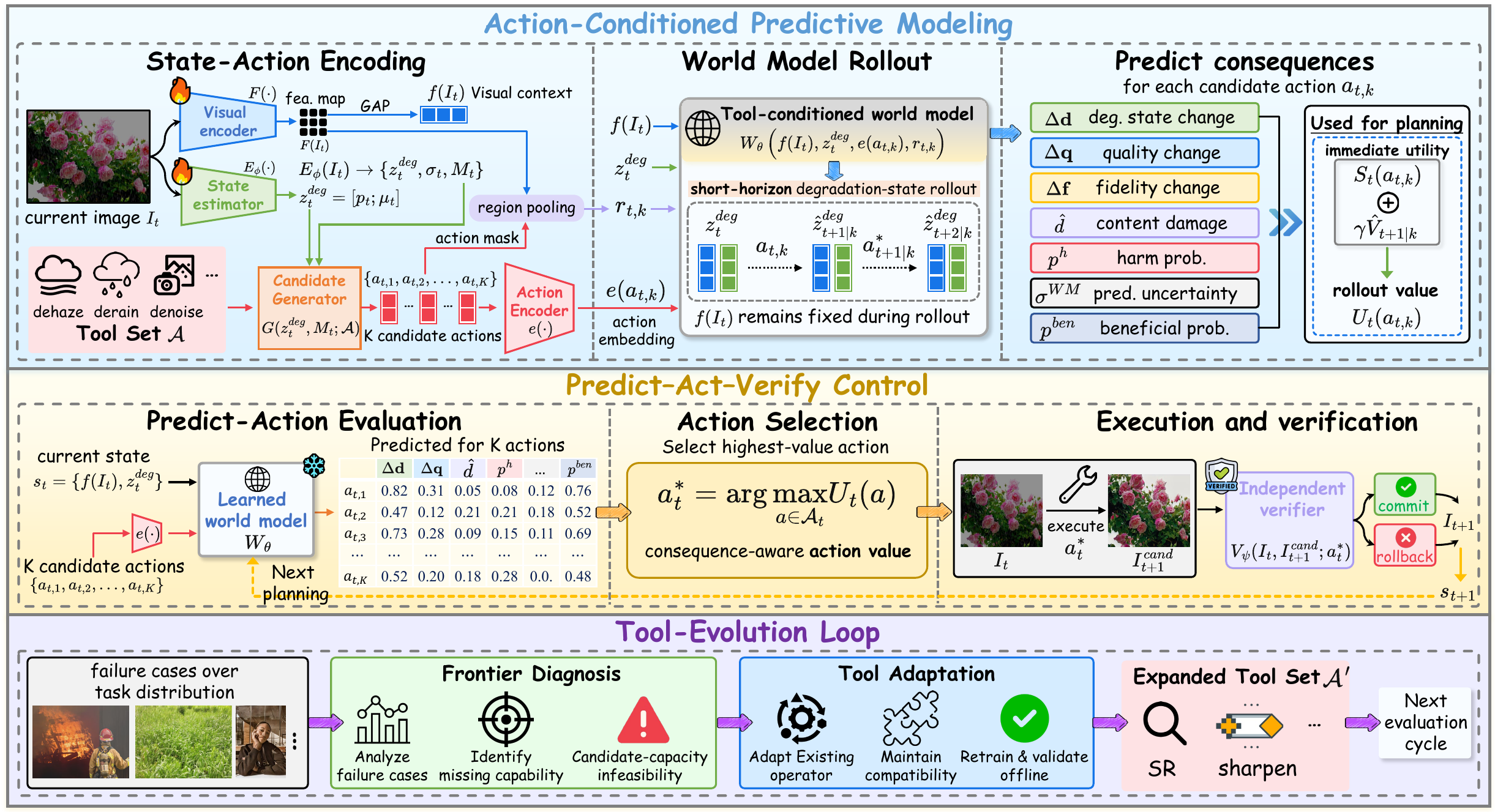}

\vspace{-1em}
    \caption{Overview of \method. The world model predicts action consequences
    before execution, while the predict--act--verify controller executes and
    verifies the selected action with commit or rollback. A tool-evolution loop
    performs candidate-frontier audit and targeted tool adaptation.}
    \label{fig:overview}

\vspace{-1em}
\end{figure}

\subsection{Overview}

\method{} follows a predict--act--verify loop (Fig.~\ref{fig:overview}). A
tool-conditioned world model predicts structured action consequences before
execution (Sec.~\ref{sec:world-model}), and an independent verifier commits or
rolls back each executed action (Sec.~\ref{sec:pav}). An offline
candidate-frontier audit then checks whether the candidate space limits
performance, and targeted tool adaptation is studied as a controlled way to
expand it (Sec.~\ref{sec:frontier-method}).

\subsection{Tool-Conditioned World Model}
\label{sec:world-model}

For $D$ degradation types, the degradation estimator predicts
\begin{equation}
\{
\mathbf p_t,\boldsymbol\mu_t,
\boldsymbol\sigma_t^{\rm est},\mathbf M_t
\}
=
E_\phi(I_t),
\end{equation}
where $\mathbf p_t$, $\boldsymbol\mu_t$, and
$\boldsymbol\sigma_t^{\rm est}$ denote degradation presence, severity, and
estimation uncertainty, respectively, and $\mathbf M_t$ holds spatial
degradation maps. The compact degradation state is
$\mathbf z_t^{\rm deg}=[{\mathbf p}_t;{\boldsymbol\mu}_t]$.

For each candidate action $a_{t,k}$, $k=1,\ldots,K$, its spatial mask
$\mathbf M_{t,k}$ is used for action-conditioned region pooling to obtain the regional
feature $\mathbf r_{t,k}$ (the OOF experiments in
Secs.~\ref{sec:openloop-ranking}--\ref{sec:closedloop} use global masks;
Appendix~\ref{sec:supp-world-arch}), and $e(a_{t,k})$ encodes the action. The
world model predicts
\begin{equation}
W_\theta
\big(
f(I_t),\mathbf r_{t,k},\mathbf z_t^{\rm deg},e(a_{t,k})
\big)
=
\left\{
\widehat{\Delta\mathbf d}_{t,k},
\widehat{\Delta\mathbf q}_{t,k},
\widehat{\Delta\mathbf f}_{t,k},
\hat d_{t,k},
p_{t,k}^{\rm harm},
p_{t,k}^{\rm sev},
p_{t,k}^{\rm ben},
\boldsymbol\sigma_{t,k}^{\rm WM}
\right\},
\label{eq:world-model}
\end{equation}
where $\widehat{\Delta\mathbf d}_{t,k}$ is the predicted degradation reduction
($\Delta d_j=\mu_j^{\rm before}-\mu_j^{\rm after}$), $\widehat{\Delta\mathbf q}_{t,k}$
and $\widehat{\Delta\mathbf f}_{t,k}$ are changes in proxy quality and
Y-channel fidelity, $\hat d_{t,k}$ denotes content damage,
$p_{t,k}^{\rm harm}$, $p_{t,k}^{\rm sev}$, and $p_{t,k}^{\rm ben}$ denote harm,
severe-harm, and benefit probabilities, and $\boldsymbol\sigma_{t,k}^{\rm WM}$
denotes predictive uncertainty.

The same predictions can drive a short-horizon rollout. Starting from
the imagined state $\widehat{\mathbf z}^{(0)}=\mathbf z_t^{\rm deg}$, an imagined
action updates the residual degradation severity as
\begin{equation}
\widehat{\boldsymbol\mu}^{(h+1)}
=
\operatorname{clip}_{[0,1]}
\left(
\widehat{\boldsymbol\mu}^{(h)}
-
\mathbb E_\theta
\left[
\Delta\mathbf d
\mid
f(I_t),
\widehat{\mathbf z}^{(h)},
a^{(h)}
\right]
\right).
\label{eq:degradation-rollout}
\end{equation}
Here, $a^{(h)}$ denotes the full action conditioning, including its encoded
attributes and regional feature; the visual features of $I_t$ and the
presence estimate $\mathbf p_t$ stay fixed, and only the severity is propagated.

\paragraph{World-model learning.}
The world model is trained on observed before--after tool transitions,
supervised for degradation, quality, and fidelity changes, content
damage, and beneficial, harmful, and severely harmful outcomes. Continuous heads use
Gaussian negative log-likelihood, binary heads use cross-entropy,
and pairwise ranking losses on a proxy utility and on the fidelity gain
encourage correct within-state action ordering:
\begin{equation}
\mathcal L_W =
\textstyle\sum_{c\in\{\Delta d,\Delta q,d\}}\beta_c\mathcal L_c
+\sum_{b\in\{{\rm ben},{\rm harm},{\rm sev}\}}\beta_b\mathcal L_b
+\beta_f\mathcal L_{\Delta f}
+\beta_{\rm rank}^{f}\mathcal L_{\rm rank}^{f}
+\beta_{\rm rank}\mathcal L_{\rm rank}.
\label{eq:world-loss}
\end{equation}
Supervision construction, architecture, and loss weights are detailed in
Appendix~\ref{sec:supp-method}.

\subsection{Predict--Act--Verify Planning}
\label{sec:pav}

For each candidate action $a$ targeting degradation $\ell$, the world model
yields target-degradation gain $g$, proxy-quality gain $q$, fidelity gain $v$,
non-target side effect $\xi$, content damage $d$, predictive uncertainty $u$,
and benefit probability $p^{\rm ben}$, forming a planning score $S_t$ and
structured utility $U_t$:
\begin{equation}
\begin{aligned}
S_t(a)={}&
\lambda_g g+\lambda_q q+\lambda_v v
-\lambda_\xi \xi-\lambda_d d-\lambda_u u-\lambda_c C(a),\\
U_t(a)={}&
p^{\rm ben}S_t(a)-(1-p^{\rm ben})\rho ,
\end{aligned}
\label{eq:action-value}
\end{equation}
where
$v=\widehat{\Delta\mathrm{PSNR}}_Y+10\,\widehat{\Delta\mathrm{SSIM}}_Y$
predicts the fidelity gain
$\Delta F_Y=\Delta\mathrm{PSNR}_Y+10\,\Delta\mathrm{SSIM}_Y$,
$\xi=\sum_{j\neq \ell}[-\widehat{\Delta d}_j]_+$ measures predicted degradation
increases outside the target, $C(a)$ is a relative cost prior, and $\rho$ is the cost of a rejected
call.
The planner selects
$a_t^*=\arg\max_{a\in\mathcal A_t}U_t(a)$.
Optional variants add harm penalties or an $H=2$ lookahead via
Eq.~\eqref{eq:degradation-rollout}, executing only the first action
(Appendix~\ref{sec:supp-risk}); neither improved consistently in our
ablation (Table~\ref{tab:policy-ablation}), so Secs.~\ref{sec:openloop-ranking}--\ref{sec:closedloop} use the
one-step score.

\begin{figure}[t]
    \centering
    \includegraphics[width=.95\linewidth]{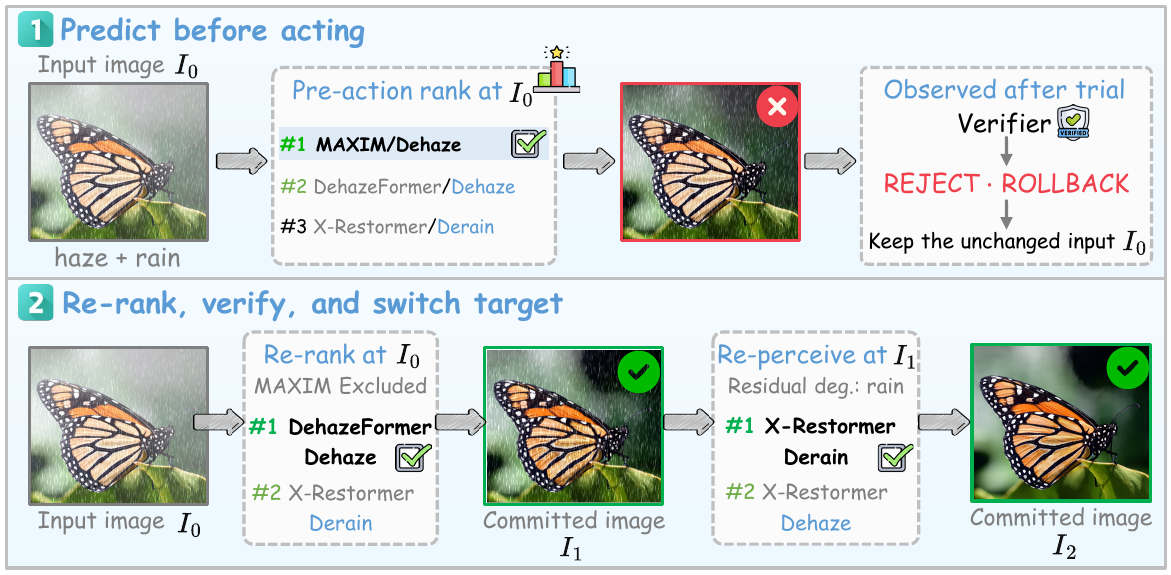}
\vspace{-1em}
    \caption{An illustrative predict--act--verify trajectory under haze and rain.}
    \label{fig:trace}
\vspace{-1em}
\end{figure}

After execution, a separately trained verifier $V_\psi$ evaluates the observed
transition (Appendix~\ref{sec:supp-verifier}): verified candidates are committed,
others rolled back, and the state is re-estimated from the observed image before
the agent replans. Fig.~\ref{fig:trace} shows an example: the
top-ranked MAXIM dehazing action is rejected and rolled back; re-ranking without it commits
DehazeFormer~\citep{song2023dehazeformer}; the re-estimated $I_1$ has only
rain left, which X-Restormer~\citep{chen2024xrestormer} removes.

\subsection{Candidate-Frontier Audit and Targeted Tool Adaptation}
\label{sec:frontier-method}

No controller can outperform the best candidates reachable from the tool
set $\mathcal A$, so we audit this candidate space offline. Let $\mathcal C_i(\mathcal A)$ denote the frozen candidate outputs
reachable for state $i=1,\ldots,N$. The audit selects one candidate per
state to maximize a dataset-level fidelity metric $m_F$ subject to floors on
other metrics:
\begin{equation}
m_F^\star(\boldsymbol\tau)=
\max_{c_i\in\mathcal C_i(\mathcal A)}\;
\frac{1}{N}\sum_{i=1}^{N}m_F(c_i)
\quad
\text{s.t.}\quad
\frac{1}{N}\sum_{i=1}^{N}m_r(c_i)\ge\tau_r,
\quad \forall r\in\mathcal M ,
\label{eq:frontier-audit}
\end{equation}
where $\tau_r$ is the floor on metric $m_r$. Sweeping $\boldsymbol\tau$ traces
the attainable frontier of the frozen pool, and a set of targets is jointly
attainable if and only if it lies on or below this frontier. Reference images
are used only for this post-hoc analysis and are never available to the
deployment-time controller.

To test whether extending the tool set can improve the attainable operating
point, we adapt an existing restoration operator on external training pairs constructed with task-matched degradation processes and add it to the tool set,
$\mathcal A'=\mathcal A\cup\{a_{\rm adapt}\}$.
Adaptation and validation are performed offline without benchmark references;
we then re-audit the expanded pool and incorporate the adapted operator into
the deployed system (Sec.~\ref{sec:frontier}).

\vspace{-1em}
\section{Experiments}
\vspace{-1em}

\subsection{Datasets and Experimental Setup}
\label{sec:experiments}

\method is evaluated on MiO100~\citep{mioir2024} under the AgenticIR
protocol~\citep{agenticir2025}, with 160 official compound-degradation states from
20 clean-image identities for training and 1,440 test images across
Groups~A--C. We use five-fold cross-fitting with disjoint clean-image
identities between fitting and outer validation. The
base tool set $\mathcal A$ contains 33 restoration actions covering eight
degradation types. Sec.~\ref{sec:main-results} (Tables~\ref{tab:main}--\ref{tab:realworld}) reports the full system, in
which the world model pre-ranks root actions, only high-value roots are
executed and expanded, and cross-fitted candidate-value models select the
output, with a fixed output blend for Groups~A and~B (Appendix~\ref{sec:supp-candidate-value}).
For Group~C, it also includes a HAT-S operator~\citep{chen2023hat} adapted on
4,000 external degradation-matched pairs, without MiO100 HQ images or test
references (Appendix~\ref{sec:supp-tool-evolution}). Controlled planner
comparisons (Secs.~\ref{sec:openloop-ranking}--\ref{sec:closedloop}) use
matched candidate pools and tool sets.

Metrics are PSNR, SSIM~\citep{wang2004ssim}, LPIPS~\citep{zhang2018lpips},
MANIQA~\citep{yang2022maniqa}, CLIP-IQA~\citep{wang2023clipiqa}, and
MUSIQ~\citep{ke2021musiq}; PSNR and SSIM are computed on the Y channel of
YCbCr.

\begin{table}[t]
\caption{Quantitative comparison on the three compound-degradation subsets of MiO100.
Best and second-best results are marked in bold and underlined, respectively.}
\tablecaptiongap
\label{tab:main}
\centering
\small
\setlength{\tabcolsep}{3pt}
\renewcommand{\arraystretch}{0.9}

\begin{tabular}{cclcccccc}
\toprule
Group & Type & Method &
PSNR$\uparrow$ & SSIM$\uparrow$ & LPIPS$\downarrow$ &
MANIQA$\uparrow$ & CLIP-IQA$\uparrow$ & MUSIQ$\uparrow$ \\
\midrule

\multirow{11}{*}{A}
& \multirow{6}{*}{All-in-One}
& \mv{AirNet}{CVPR'22}
& 19.13 & 0.602 & 0.428 & 0.258 & 0.393 & 42.46 \\

&& \mv{PromptIR}{NeurIPS'23}
& 20.06 & 0.609 & 0.413 & 0.263 & 0.401 & 42.62 \\

&& \mv{MiOIR}{arXiv'24}
& 20.84 & 0.656 & 0.372 & 0.245 & 0.393 & 47.82 \\

&& \mv{DA-CLIP}{ICLR'24}
& 19.58 & 0.603 & 0.427 & 0.242 & 0.414 & 42.51 \\

&& \mv{InstructIR}{ECCV'24}
& 18.03 & 0.575 & 0.443 & 0.266 & 0.353 & 45.77 \\

&& \mv{AutoDIR}{ECCV'24}
& 19.64 & 0.629 & 0.397 & 0.250 & 0.377 & 47.01 \\

\cmidrule(lr){2-9}

& \multirow{5}{*}{Agent}
& \mv{AgenticIR}{ICLR'25}
& 21.04 & 0.682 & 0.315 & 0.307 & 0.447 & 56.88 \\

&& \mv{MAIR}{IJCV'26}
& 21.02 & 0.672 & \underline{0.296} & 0.333 & 0.475 & 59.19 \\

&& \mv{4KAgent}{NeurIPS'25}
& 21.48 & 0.672 & 0.302 & \underline{0.375} &
\textbf{0.554} & \textbf{63.19} \\

&& \mv{SEAR}{ECCV'26}
& \underline{21.80} & \underline{0.696} & 0.302 &
0.341 & 0.504 & 61.29 \\

\rowcolor{oursblue}
&& \method
& \textbf{22.61} & \textbf{0.717} & \textbf{0.292} &
\textbf{0.383} & \underline{0.540} & \underline{62.78} \\

\midrule

\multirow{11}{*}{B}
& \multirow{6}{*}{All-in-One}
& \mv{AirNet}{CVPR'22}
& 19.31 & 0.657 & 0.367 & 0.288 & 0.427 & 47.88 \\

&& \mv{PromptIR}{NeurIPS'23}
& 20.47 & 0.670 & 0.337 & 0.289 & 0.429 & 48.10 \\

&& \mv{MiOIR}{arXiv'24}
& 20.56 & 0.691 & 0.324 & 0.264 & 0.433 & 51.87 \\

&& \mv{DA-CLIP}{ICLR'24}
& 18.56 & 0.595 & 0.441 & 0.244 & 0.415 & 43.70 \\

&& \mv{InstructIR}{ECCV'24}
& 18.34 & 0.624 & 0.407 & 0.302 & 0.379 & 50.94 \\

&& \mv{AutoDIR}{ECCV'24}
& 19.90 & 0.664 & 0.354 & 0.253 & 0.399 & 49.64 \\

\cmidrule(lr){2-9}

& \multirow{5}{*}{Agent}
& \mv{AgenticIR}{ICLR'25}
& 20.55 & 0.701 & 0.307 & 0.320 & 0.465 & 57.57 \\

&& \mv{MAIR}{IJCV'26}
& 20.92 & 0.700 & \textbf{0.279} & 0.354 & 0.508 & 60.98 \\

&& \mv{4KAgent}{NeurIPS'25}
& 20.95 & 0.673 & 0.302 & \underline{0.373} &
\underline{0.551} & \underline{62.69} \\

&& \mv{SEAR}{ECCV'26}
& \textbf{22.13} & \underline{0.725} & 0.289 &
0.345 & 0.513 & 60.65 \\

\rowcolor{oursblue}
&& \method
& \underline{22.00} & \textbf{0.730} & \underline{0.280} &
\textbf{0.388} & \textbf{0.555} & \textbf{63.31} \\

\midrule

\multirow{11}{*}{C}
& \multirow{6}{*}{All-in-One}
& \mv{AirNet}{CVPR'22}
& 17.95 & 0.515 & 0.578 & 0.185 & 0.311 & 30.12 \\

&& \mv{PromptIR}{NeurIPS'23}
& 18.51 & 0.517 & 0.576 & 0.191 & 0.310 & 29.71 \\

&& \mv{MiOIR}{arXiv'24}
& 15.63 & 0.490 & 0.538 & 0.172 & 0.289 & 37.95 \\

&& \mv{DA-CLIP}{ICLR'24}
& 18.53 & 0.532 & 0.534 & 0.192 & 0.348 & 33.87 \\

&& \mv{InstructIR}{ECCV'24}
& 17.09 & 0.514 & 0.558 & 0.173 & 0.254 & 33.69 \\

&& \mv{AutoDIR}{ECCV'24}
& 18.61 & 0.544 & 0.502 & 0.205 & 0.294 & 37.86 \\

\cmidrule(lr){2-9}

& \multirow{5}{*}{Agent}
& \mv{AgenticIR}{ICLR'25}
& 18.82 & 0.547 & 0.449 & 0.270 & 0.395 & 48.68 \\

&& \mv{MAIR}{IJCV'26}
& 19.42 & 0.554 & \textbf{0.414} & 0.280 & 0.424 & 51.36 \\

&& \mv{4KAgent}{NeurIPS'25}
& 19.77 & 0.563 & 0.427 & \textbf{0.355} &
\textbf{0.523} & \textbf{55.56} \\

&& \mv{SEAR}{ECCV'26}
& \underline{20.61} & \underline{0.593} & \underline{0.417} &
0.274 & 0.442 & 52.73 \\

\rowcolor{oursblue}
&& \method
& \textbf{21.05} & \textbf{0.604} & 0.418 &
\underline{0.332} & \underline{0.505} & \underline{54.78} \\

\bottomrule

\vspace{-3em}

\end{tabular}
\end{table}

\begin{table}[t]
\caption{Quantitative comparison on the paired real-world 100-image set used
in MAIR.}
\tablecaptiongap
\label{tab:realworld}
\centering
\renewcommand{\arraystretch}{0.9}
\resizebox{0.8\linewidth}{!}
{%
\begin{tabular}{lcccccc}
\toprule
Method & PSNR$\uparrow$ & SSIM$\uparrow$ & LPIPS$\downarrow$
& MANIQA$\uparrow$ & CLIP-IQA$\uparrow$ & MUSIQ$\uparrow$ \\
\midrule
AirNet   & 23.31 & \underline{0.747} & 0.448 & 0.236 & 0.327 & 35.27 \\
PromptIR & 23.86 & \textbf{0.754} & 0.434 & 0.241 & 0.327 & 35.48 \\
MiOIR    & \underline{24.00} & \underline{0.747} & 0.424 & 0.239 & 0.351 & 36.85 \\
DA-CLIP  & 23.73 & 0.744 & 0.427 & 0.251 & 0.333 & 36.41 \\
AutoDIR  & 20.75 & 0.664 & 0.479 & 0.247 & 0.315 & 42.68 \\
\midrule
AgenticIR & 23.93 & 0.721 & 0.373 & 0.315 & 0.451 & 52.91 \\
4KAgent   & 23.63 & 0.724 & \underline{0.357} & \underline{0.320}
          & 0.451 & 54.46 \\
SEAR      & \textbf{24.41} & 0.743 & \textbf{0.337}
          & 0.317 & \underline{0.452} & \underline{54.67} \\

\rowcolor{oursblue}
\method   & 22.64 & 0.731 & 0.370
          & \textbf{0.389} & \textbf{0.525} & \textbf{57.33} \\
\bottomrule
\end{tabular}}
\vspace{-1em}
\end{table}

\subsection{MiO100 Main Results}
\label{sec:main-results}

Table~\ref{tab:main} compares \method with representative all-in-one and
agentic methods on MiO100. \method achieves the highest SSIM on all
three groups and the highest PSNR on Groups~A and~C.
On Group~A it also leads in LPIPS and MANIQA
(22.61\,dB, 0.292, and 0.383; SSIM 0.717), and on Group~B it leads in MANIQA,
CLIP-IQA, and MUSIQ. On Group~C it leads in PSNR and SSIM
(21.05\,dB and 0.604) but trails 4KAgent on the three perceptual metrics. Qualitative comparisons in
Fig.~\ref{fig:qualitative_mio100} show similar trends (more examples in
Appendix~\ref{app:qualitative}).

\begin{figure}[t]
    \centering
    \includegraphics[width=\linewidth]{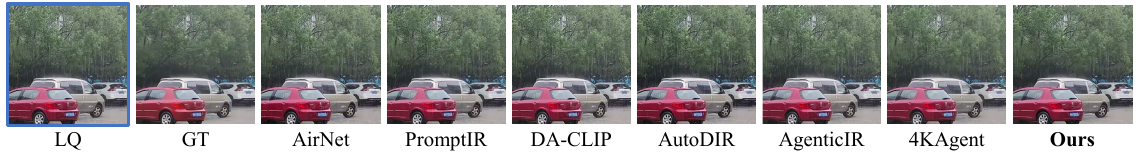}
\vspace{-2em}
    \caption{Qualitative comparison on a representative example from the paired
real-world dataset.}
    \label{fig:qualitative_real}
\vspace{-1.5em}
\end{figure}

On the paired real-world set (Table~\ref{tab:realworld}), \method leads on
MANIQA, CLIP-IQA, and MUSIQ
(0.389, 0.525, and 57.33), but its PSNR (22.64\,dB) falls behind all other
agents (SEAR reaches 24.41\,dB), pointing to a fidelity--perception trade-off
on this set that deserves further investigation. Qualitative results are in
Fig.~\ref{fig:qualitative_real}.

\begin{figure}[t]
    \centering
    \includegraphics[width=\linewidth]{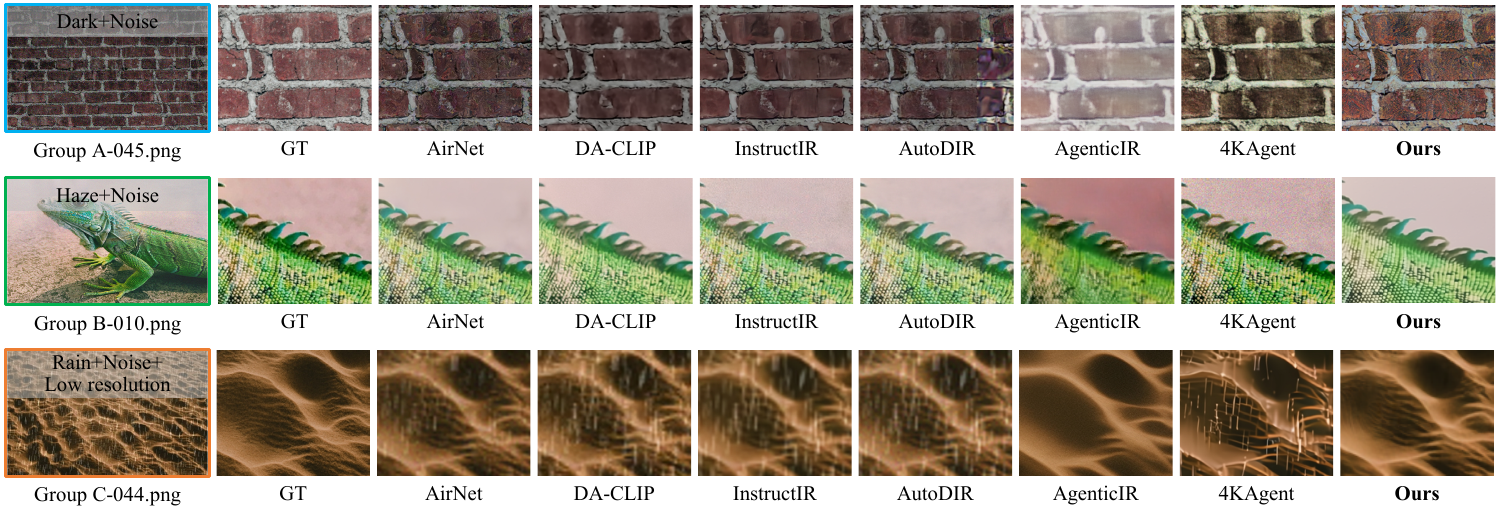}
\vspace{-2em}
    \caption{Qualitative comparison on representative examples from MiO100 Groups~A--C.}
\vspace{-1em}
    \label{fig:qualitative_mio100}
\end{figure}

\subsection{Open-Loop Pre-Execution Ranking}
\label{sec:openloop-ranking}

This section evaluates the world model before any tool is executed. On 160
clean-identity-disjoint out-of-fold (OOF) states, each method scores the same
ranker-independent pool of 2,147 executed candidates and selects one action
without verifier, retry, or lookahead. We evaluate top-1 headroom recovery
(HR), harmful selections, within-state ranking, and harm discrimination. HR is
the ratio of the mean realized $\Delta F_Y$ (Sec.~\ref{sec:pav}) of the
selected action to that of the per-state best action. All CIs are 95\%
clean-identity cluster-bootstrap intervals of paired differences.

\begin{figure*}[t]
\centering
\includegraphics[width=1\linewidth]{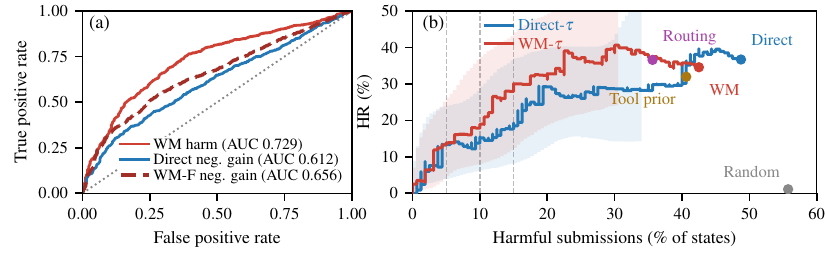}

\vspace{-2em}
\caption{Risk information from pre-execution prediction; brackets and bands
are 95\% clean-identity cluster-bootstrap intervals.
(a) ROC for detecting harmful actions among 2,147 candidates. Paired AUC gains
of $p^{\rm harm}$: 0.118
$[0.091, 0.146]$ over Direct and 0.073 $[0.054, 0.093]$ over WM-F.
(b) HR versus harmful submissions when the top action may be replaced by a
no-op. Lines sweep an abstention threshold on predicted harm (WM-$\tau$) or
gain (Direct-$\tau$), with bands shown where all draws have support; dots:
top-1 results of Table~\ref{tab:sec43_openloop}; dashed lines: 5/10/15\% harm.}
\label{fig:sec43_openloop}

\vspace{-1em}
\end{figure*}

The world model's main advantage is not higher one-shot fidelity, but richer
pre-execution consequence information. Its HR is similar to Direct
(34.6\% vs.\ 36.7\%; difference $-2.1$ points, CI $[-15.5, 11.8]$), while it
better preserves within-state candidate ordering
($r_s=0.309$ vs.\ 0.247; difference $0.062$, CI $[0.009, 0.120]$).
More clearly, its harm probability distinguishes harmful from non-harmful
candidates better than Direct or WM-F (ROC-AUC 0.729 vs.\ 0.612 and 0.656;
Fig.~\ref{fig:sec43_openloop}a). At matched harm rates of 5, 10, and 15\%,
harm-based abstention recovers $+0.0$, $+3.5$, and $+10.7$ points more HR than
gain-based abstention, although all CIs include zero
(Fig.~\ref{fig:sec43_openloop}b). Overall, the world model contributes most
through candidate ordering and risk information rather than top-1 fidelity
prediction alone. Appendix~\ref{app:sec43} gives the full paired comparisons.

\begin{table}[t]
\caption{Open-loop top-1 selection on 160 OOF states with 2,147 executed
candidates. HR: share of oracle headroom recovered. Harm: the selected
action lowers Y-PSNR by more than 0.05\,dB, Y-SSIM by more than 0.005, or
mean proxy quality. Severe: Y-PSNR drop $\ge$1\,dB or Y-SSIM
drop $\ge$0.05. $r_{\rm s}$: mean within-state Spearman correlation of scores
with realized $\Delta F_Y$. WM-F: fidelity head alone.}
\label{tab:sec43_openloop}
\centering\small
\setlength{\tabcolsep}{4pt}
\begin{tabular}{lrrrrrrr}
\toprule
Method & HR (\%)$\uparrow$ & Harm (\%)$\downarrow$ & Severe (\%)$\downarrow$
& $\Delta$PSNR$_Y$$\uparrow$ & $\Delta$SSIM$_Y$$\uparrow$
& $\Delta$LPIPS$\downarrow$ & $r_{\rm s}\uparrow$ \\
\midrule
\textit{Random}     &   1.1 & 55.7 & 28.2 & $-$0.10 & 0.015 & $-$0.038 & -- \\
Tool prior          &  31.9 & 40.6 & 13.8 &    0.98 & 0.058 & $-$0.067 & 0.284 \\
Degradation routing &  36.6 & 35.6 & 15.0 &    1.27 & 0.052 & $-$0.079 & 0.378 \\
Direct scorer       &  36.7 & 48.8 & 27.5 &    1.26 & 0.053 & $-$0.056 & 0.247 \\

World model (WM)    &  34.6 & 42.5 & 21.9 &    1.14 & 0.055 & $-$0.069 & 0.309 \\
WM-F                &  31.9 & 46.2 & 23.1 &    1.05 & 0.051 & $-$0.057 & 0.314 \\
\midrule
\textit{Oracle}     & 100.0 &  5.0 &  0.0 &    3.86 & 0.102 & $-$0.116 & -- \\
\bottomrule
\end{tabular}
\vspace{-1em}
\end{table}

\begin{table}[t]
\caption{Matched closed-loop comparison on the 160 OOF states with a shared
candidate pool, verifier, and budget $B=4$. Calls and rejects are per image;
time counts restoration-tool execution only. Rejects, \method minus reactive:
$-1.49$ per image, CI $[-1.78,-1.21]$.}
\label{tab:matched-reactive}
\centering\small
\setlength{\tabcolsep}{5pt}
\begin{tabular}{lrrrrrrr}
\toprule
Policy & Calls$\downarrow$ & Rejects$\downarrow$ & Time (s)$\downarrow$
& PSNR$_Y$$\uparrow$ & SSIM$_Y$$\uparrow$ & LPIPS$\downarrow$
& Harm (\%)$\downarrow$ \\
\midrule
Reactive trial-and-error & 2.97 & 2.36 & 39.6 & 21.16 & 0.676 & 0.422 & 15.6 \\
Fidelity-head ranking    & 1.75 & 0.90 & 25.8 & 22.27 & 0.689 & 0.420 & 18.1 \\
\method (structured)     & 1.71 & 0.86 & 25.3 & 22.08 & 0.690 & 0.416 & 14.4 \\
\bottomrule
\end{tabular}
\vspace{-1.5em}
\end{table}

\subsection{Predict-before-Act versus Reactive Trial-and-Error}
\label{sec:closedloop}

The previous section evaluated ranking from frozen outcomes; here we test
whether better ranking saves real tool calls when failures
are costly. In a controlled replay on the same 160 OOF states, three policies
share the candidate pool, an annotated degradation state, the post-execution
verifier, and a budget of four real calls; each executes candidates in its
own order, rolls back rejected outputs, and stops at the first verified one.
The pool contains the 1,415 candidates executed when constructing the training
transitions: all tools for each state's annotated degradation types, plus up
to two randomly drawn tools from other types (7--12 per state). Because the shared
state uses annotated degradation types and severities computed
from the clean reference, the replay isolates execution order under an
accurate state rather than estimating deployment performance
(Sec.~\ref{sec:openloop-ranking}). Reactive
trial-and-error orders targets by presence times severity (floored at 0.05) and tools by a
per-target route fitted on two inner identities; fidelity-head ranking (WM-F) and
\method order the same candidates by predicted fidelity gain $v$ and by
structured utility $U_t$, respectively. With the verifier threshold fixed per fold,
only the ordering differs.

Pre-execution ranking reduces wasted calls by a large margin
(Table~\ref{tab:matched-reactive}). Against reactive trial-and-error,
\method uses 42.5\% fewer real calls (paired difference $-1.26$ per image, CI
$[-1.50,-1.02]$) and 63.4\% fewer rejected calls, and improves Y-PSNR by
0.92\,dB (CI $[0.49,1.34]$), partly because it more often finds a verified
candidate within budget (no-commit rate 23.1 points lower, CI $[15.6, 30.6]$).
Fidelity-head
ranking achieves similar savings, so the efficiency gain comes mainly from
ranking candidates before execution rather than from the structured utility
specifically. Against fidelity-head ranking, the structured
utility does reduce harm by 3.75 points (CI $[0.6, 7.5]$),
consistent with the open-loop ablation of
$p^{\rm ben}$. Other
paired intervals and budget sensitivity are in Appendix~\ref{app:closedloop}.

\begin{figure}[t]
    \centering
    \includegraphics[width=.98\linewidth]{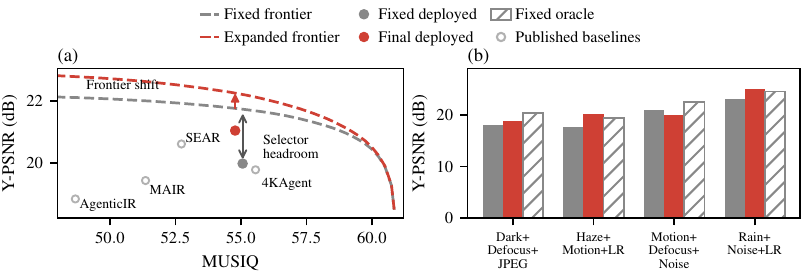}
\caption{Group~C frontier audit (post hoc, using test references).
(a) Attainable mean Y-PSNR vs.\ MUSIQ for the fixed and expanded pools
(dashed); gray: deployed fixed-pool system; red: final system, built on a
different fixed-pool configuration (Sec.~\ref{sec:frontier}); hollow:
baselines of Table~\ref{tab:main}. (b) Per-combination Y-PSNR (LR: low
resolution); oracle at the fixed system's MUSIQ level.}
\label{fig:groupc-frontier}
\vspace{-1em}
\end{figure}

\subsection{Candidate-Frontier Audit and Targeted Tool Adaptation}
\label{sec:frontier}

No amount of better prediction helps if no reachable candidate is good enough.
We therefore audit whether Group~C is limited by selection or by what the
33-action tool set can produce. Using test references post hoc, we score all 21,385
frozen Group~C candidates and solve an exact one-candidate-per-state
mixed-integer program that maximizes mean Y-PSNR subject to a floor on mean
MUSIQ; sweeping the floor traces the attainable fidelity--perception frontier
of the fixed pool (Fig.~\ref{fig:groupc-frontier}a). This oracle is not
deployable, but it shows where the ceiling is. The deployed fixed-pool
configuration lies well inside the frontier: at its MUSIQ level, a
reference-assisted selection from the same pool reaches 21.74
versus 19.97\,dB Y-PSNR, leaving 1.77\,dB
of selector headroom. Adapting HAT-S on 4,000 external replay pairs
expands the pool for the 200 low-resolution states and raises the attainable
frontier by 0.50\,dB at the final MUSIQ level. Our final Group~C system builds
on a different fixed-pool configuration and fuses the adapted tool's output
into the low-resolution states; it reaches 21.05/0.604 Y-PSNR/Y-SSIM at MUSIQ
54.78 but still lies 0.7\,dB below the fixed-pool frontier. On the two
low-resolution combinations it exceeds the fixed-pool oracle, while the
other two retain 1.6--2.6\,dB of selector headroom
(Fig.~\ref{fig:groupc-frontier}b). Tool adaptation is one practical route to a better operating point, but closing the selector
headroom offers a larger gain.

\section{Conclusion}

We presented PreActIR, an agentic restoration framework that looks before it
restores: a tool-conditioned world model predicts the consequences of each
candidate tool before execution, and an independent verifier commits or rolls
back the executed one. In controlled replays with matched candidates, verifier,
and budget, predicting before acting substantially reduces real and rejected
tool calls while improving restoration quality, and on MiO100 the full system
achieves the best overall performance on all three compound-degradation groups. 
The candidate-frontier audit shows where the remaining gains lie: the
existing tool set holds considerable headroom for better selection, and
targeted adaptation can push the frontier further. Two directions remain
open: validating these gains under estimated degradation states and
establishing whether multi-step imagination adds value beyond single-step
ranking.

\section*{AI Use Statement}
In this work, generative AI tools were used to edit the manuscript and improve readability and language. All AI-suggested edits were reviewed against the original text by the authors, who take full responsibility for the final content, including all text, claims, and AI-assisted artifacts.
\bibliographystyle{plainnat}
\bibliography{preactir_iclr2027}

\clearpage
\appendix

\FloatBarrier
\section{Additional Method Details}
\label{sec:supp-method}

\subsection{World-Model Architecture}
\label{sec:supp-world-arch}

The world model receives a $256\times256$ image, the degradation-state
estimates for the $D=8$ degradation types, and one of the $|\mathcal A|=33$
tool actions. The image encoder contains four stages with channel dimensions
$[32,64,128,256]$. The first stage uses a $5\times5$ stride-2
convolution, while the remaining stages use $3\times3$ stride-2
convolutions. Each stage is followed by a two-layer residual block
with GroupNorm and SiLU activations.

The deepest visual features are aggregated using global-average pooling
and action-mask-weighted pooling, and both are projected to 256
dimensions. They are concatenated with 16-dimensional
presence/severity features, a 48-dimensional tool embedding, a
24-dimensional target-degradation embedding, and three scalar inputs
for action strength, mask area, and relative compute cost. The resulting
603-dimensional representation is processed by two width-384
Linear--LayerNorm--SiLU--Dropout blocks with dropout 0.15.
Separate heads on these shared features output Gaussian means and log
variances for the eight-dimensional degradation change $\Delta\mathbf d$, the
four-dimensional proxy-quality change $\Delta\mathbf q$, the two-dimensional
Y-channel fidelity change $\Delta\mathbf f$, and the scalar content damage
$d$, together with binary logits for beneficial, harmful, and severely harmful
outcomes.
The damage mean is passed through a sigmoid, and all predicted log variances
are clipped to $[-6,2]$.

The action mask, region area, and strength are retained as a general
action interface. The OOF experiments in
Secs.~\ref{sec:openloop-ranking}--\ref{sec:closedloop}, however, use global
actions with fixed strength $\omega=1$. We therefore do not claim
independently validated regional control or strength generalization in the
current experiments.

\subsection{Supervision Label Construction}
\label{sec:supp-labels}

Training data are generated from real restoration-tool interventions,
retaining the image before and after each action. Degradation labels follow
the sign convention of Sec.~\ref{sec:world-model}:
$\Delta d_j=\mu_j^{\rm before}-\mu_j^{\rm after}$, so positive values denote
improvement. A general action is
represented as
\begin{equation}
    a=(m,\ell,\omega,\mathbf M_a),
\end{equation}
where $m$ denotes a specific restoration tool, $\ell$ its compatible
target degradation, $\omega$ the action strength, and $\mathbf M_a$ the affected
region. Tool--target pairs are restricted to compatible combinations.

For a reference image $I^*$, we define pixel and gradient errors as
\begin{equation}
    \mathrm{err}(I)=\operatorname{mean}_c |I-I^*|,
    \qquad
    \mathrm{err}_\nabla(I)=|\nabla I-\nabla I^*|.
\end{equation}
The content-damage label, measured outside the action region, is
\begin{equation}
d=
\operatorname{clip}_{[0,1]}
\left(
\mathbb E_{1-\mathbf M_a}[\mathrm{err}(I')-\mathrm{err}(I)]_+
+
\frac{1}{2}
\mathbb E_{1-\mathbf M_a}[\mathrm{err}_\nabla(I')-\mathrm{err}_\nabla(I)]_+
\right),
\label{eq:supp-damage}
\end{equation}
where $I'$ denotes the image after applying the action. For a global action,
$\mathbf M_a\equiv\mathbf 1$, the region outside the action is empty and we set
$d=0$ by convention.
The four-dimensional proxy-quality change is
\begin{equation}
    \Delta\mathbf q=\mathbf q(I')-\mathbf q(I),
\end{equation}
where $\mathbf q$ is an RGB-fidelity proxy used only for constructing
training supervision. Its components are RGB-PSNR clipped at 50~dB
and divided by 50, RGB-SSIM, $1-\ell_1$ similarity, and
$\exp(-5\,\mathrm{err}_{\rm edge})$ edge similarity.
The fidelity head is supervised with a separate label, computed from the input, output, and clean reference using Y-channel PSNR and SSIM:
\begin{equation}
    \mathbf y_f=\big(\Delta\mathrm{PSNR}_Y/10,\ \Delta\mathrm{SSIM}_Y\big),
\end{equation}
so that $10\,(y_{f,1}+y_{f,2})=\Delta F_Y$. At test time, the fidelity-head
means $(\mu_{f,1},\mu_{f,2})$ are mapped back to the original units,
\begin{equation}
    \widehat{\Delta\mathrm{PSNR}}_Y=10\,\mu_{f,1},\qquad
    \widehat{\Delta\mathrm{SSIM}}_Y=\mu_{f,2},
\end{equation}
so the fidelity gain of Eq.~\eqref{eq:action-value} is
$v=\widehat{\Delta\mathrm{PSNR}}_Y+10\,\widehat{\Delta\mathrm{SSIM}}_Y
=10\,(\mu_{f,1}+\mu_{f,2})$.

For the base intervention data, an action is labeled beneficial if the
target degradation score improves by at least $0.01$, the damage score
satisfies $d\leq0.02$, and the mean four-dimensional quality change is
no smaller than $-0.025$. For the added OOF candidates, an action is
labeled beneficial if all of
\begin{equation}
\Delta\mathrm{PSNR}_{\rm RGB}\geq0.05~{\rm dB},
\qquad
\Delta\mathrm{SSIM}_{\rm RGB}\geq-0.005,
\qquad
\operatorname{mean}(\Delta\mathbf q)>0
\end{equation}
hold. The online verifier decision is retained only as an audit field and
is not used as supervision.

An action is labeled harmful ($y_{\rm harm}=1$) if any of
\begin{equation}
\Delta\mathrm{PSNR}_{Y}<-0.05~{\rm dB},
\qquad
\Delta\mathrm{SSIM}_{Y}<-0.005,
\qquad
\operatorname{mean}(\Delta\mathbf q)<0
\end{equation}
holds; the first two conditions are computed on the Y channel and the third
uses the RGB proxy above. An action that is not beneficial is therefore not
necessarily harmful; neutral transitions remain possible.
We separately define a severe event as
\begin{equation}
\Delta\mathrm{PSNR}_{Y}\leq -1~{\rm dB}
\quad\text{or}\quad
\Delta\mathrm{SSIM}_{Y}\leq -0.05.
\end{equation}
This label supervises the separate severe-harm head (Appendix~\ref{sec:supp-world-loss})
and defines the severe rates in Appendix~\ref{app:sec43}; it is not merged into
the harm label.
Because the OOF candidates are global actions, their content-damage label is
zero by this convention; risk evidence for these samples
therefore comes from the harm prediction and the post-execution verifier.

\subsection{World-Model Training Objective}
\label{sec:supp-world-loss}

For a Gaussian target $y$ with predicted mean $\mu$ and variance $\sigma^2$,
we use
\begin{equation}
\operatorname{NLL}(\mu,\log\sigma^2;y)
=
\frac{1}{2}
\left[
\log\sigma^2+\frac{(y-\mu)^2}{\sigma^2}
\right].
\end{equation}
The frozen world model (configuration V8) minimizes
\begin{equation}
\begin{split}
\mathcal L_W={}&
\beta_{\Delta d}\mathcal L_{\Delta d}
+\beta_{\Delta q}\mathcal L_{\Delta q}
+\beta_d\mathcal L_{d}
+\beta_{\rm ben}\mathcal L_{\rm ben}
+\beta_{\rm harm}\mathcal L_{\rm harm}
+\beta_{\rm sev}\mathcal L_{\rm sev}\\
&+\beta_f\mathcal L_{\Delta f}
+\beta_{\rm rank}^{f}\mathcal L_{\rm rank}^{f}
+\beta_{\rm rank}\mathcal L_{\rm rank},
\end{split}
\label{eq:supp-world-loss}
\end{equation}
where $\mathcal L_{\Delta d}$, $\mathcal L_{\Delta q}$, $\mathcal L_d$, and
$\mathcal L_{\Delta f}$ are Gaussian NLLs on $\Delta\mathbf d$,
$\Delta\mathbf q$, $d$, and the fidelity label $\mathbf y_f$
(Appendix~\ref{sec:supp-labels}), and $\mathcal L_{\rm ben}$,
$\mathcal L_{\rm harm}$, and $\mathcal L_{\rm sev}$ are binary cross-entropies
on the beneficial, harmful, and severe labels, with positive-class weights of
$1.4$ for harm and $5.0$ for severe harm. Both ranking losses compare only
actions originating from the same state: $\mathcal L_{\rm rank}^{f}$
encourages the predicted $\Delta F_Y$ to follow the within-state order of the
realized $\Delta F_Y$, and $\mathcal L_{\rm rank}$ applies a softplus pairwise
loss to signed differences of a ranking utility that combines
target-degradation improvement, $0.5$ times the mean proxy-quality
improvement, and $-2$ times the content damage, computed from both the
targets and the predictions. The weights are
\begin{equation}
\begin{aligned}
&(\beta_{\Delta d},\beta_{\Delta q},\beta_d,\beta_{\rm ben},\beta_{\rm harm},
\beta_{\rm sev},\beta_f,\beta_{\rm rank}^{f},\beta_{\rm rank})\\
&\qquad=(0.5,\,0.25,\,0.5,\,0.75,\,0.75,\,1.5,\,3.0,\,1.0,\,0.05).
\end{aligned}
\end{equation}
V8 initializes its shared trunk from the V7 world-model checkpoint, which was
trained with a different weighting and without the two fidelity terms.

\FloatBarrier
\section{Additional Planning and Verification Details}
\label{sec:supp-planning}

\subsection{Planning Score and Optional Two-Step Extension}
\label{sec:supp-risk}

For a candidate action $a$ targeting degradation $\ell$, let
$g=\widehat{\Delta d}_\ell$ denote the predicted improvement of the target
degradation, $q$ the mean of $\widehat{\Delta\mathbf q}$,
$v=10\,(\mu_{f,1}+\mu_{f,2})$ the predicted $\Delta F_Y$ obtained from the two
fidelity-head means,
\begin{equation}
\xi=\sum_{j\neq \ell}[-\widehat{\Delta d}_j]_+
\end{equation}
the side effect on non-target degradations, $d$ the predicted content
damage, and $u$ the mean predictive standard deviation across
$\widehat{\Delta\mathbf d}$, $\widehat{\Delta\mathbf q}$, and $d$. The
harm, severe-harm, and benefit probabilities are $p^{\rm harm}$,
$p^{\rm sev}$, and $p^{\rm ben}$.

The complete one-step planning score and structured utility are
\begin{equation}
\begin{split}
S_t(a)={}&
\lambda_g g+\lambda_q q+\lambda_v v
-\lambda_\xi \xi-\lambda_d d-\lambda_u u\\
&-\lambda_{\rm harm} p^{\rm harm}-\lambda_{\rm sev}p^{\rm sev}
-\lambda_c C(a),\\
U_t(a)={}&p^{\rm ben} S_t(a)-(1-p^{\rm ben})\rho .
\end{split}
\label{eq:supp-risk-score}
\end{equation}
The frozen configuration uses
\begin{equation}
(\lambda_g,\lambda_q,\lambda_v,\lambda_\xi,\lambda_d,
\lambda_u,\lambda_{\rm harm},\lambda_{\rm sev},\lambda_c)
=
(0.25,0.10,1.0,0.75,1.0,0.20,0.75,2.0,0.02),
\end{equation}
with rejection cost $\rho=0.05$. Here $p^{\rm ben}$ is the world model's
predicted probability that the action is beneficial
(Appendix~\ref{sec:supp-labels}); it is not the verifier's acceptance
probability $p^{\rm acc}$ (Appendix~\ref{sec:supp-verifier}), which is trained
and used differently (Table~\ref{tab:app_verifier}). $U_t$ is therefore a
benefit-weighted planning utility, not an expected commit value. The one-step score of
Eq.~\eqref{eq:action-value}, used in
Secs.~\ref{sec:openloop-ranking}--\ref{sec:closedloop}, sets
$\lambda_{\rm harm}=\lambda_{\rm sev}=0$; the harm-penalty variant in
Table~\ref{tab:policy-ablation} uses the values above. No calibrated risk
threshold (Appendix~\ref{sec:supp-risk-calibration}) is applied in these
experiments.

The relative compute prior $C(a)$ is manually specified and is not a
measured latency. Simple brightness operators receive values around
$0.02$--$0.05$, conventional neural restoration models around
$0.6$--$1.3$, and DiffBIR~\citep{lin2024diffbir} uses $4.0$. It acts only as a weak ranking
preference when predicted gains are similar.

The optional two-step extension (Sec.~\ref{sec:pav}) solves
\begin{equation}
\max_{a_{t:t+H-1}}
\sum_{h=0}^{H-1}\gamma^h U_{t+h}(a_{t+h}),
\label{eq:supp-mpc}
\end{equation}
with horizon $H=2$, discount factor $\gamma=0.85$, and beam size 4.
Each planning round expands only the three active degradation types
with highest confidence. The imagined second step keeps the current
visual features fixed and updates only the residual degradation state
(Eq.~\eqref{eq:degradation-rollout}). We use receding-horizon control: only
the first action of the selected sequence is executed, and the real output
is then verified, re-encoded, and used to restart planning.

Table~\ref{tab:policy-ablation} compares these options in a controlled
replay in which every policy executes exactly one real action per state.
The two-step extension does not improve over the one-step score (Y-PSNR
22.50 vs.\ 22.55\,dB; harm 17.5\% vs.\ 15.6\%), and holding the degradation
state fixed during lookahead narrows but does not close the gap. The soft
harm penalties give the highest Y-PSNR among the world-model variants
(22.56\,dB) but also the highest harm (18.8\%). We therefore use the
one-step score without penalties in
Secs.~\ref{sec:openloop-ranking}--\ref{sec:closedloop}; the settings above
document the implementation rather than support a multi-step effectiveness
claim.

\begin{table}[t]
\caption{Controlled pre-action ranking ablation on the 160 OOF states.
Every policy ranks the shared frozen pool of the closed-loop replay
(Sec.~\ref{sec:closedloop}: annotated degradation states, 1,415 cached
candidates) and executes exactly one real action per state; $H=2$ uses only
virtual lookahead. Harm rates are therefore not
comparable with Table~\ref{tab:sec43_openloop}, which uses estimated beliefs
and the 2,147-candidate pool.
$\Delta F_Y=\Delta\mathrm{PSNR}_Y+10\,\Delta\mathrm{SSIM}_Y$ is the realized
fidelity gain, and Regret is the oracle $\Delta F_Y$ minus the realized one.}
\label{tab:policy-ablation}
\centering\small
\setlength{\tabcolsep}{4pt}
\begin{tabular}{lcccccc}
\toprule
Policy & PSNR$_Y\uparrow$ & SSIM$_Y\uparrow$ & LPIPS$\downarrow$
& $\Delta F_Y\uparrow$ & Harm (\%)$\downarrow$ & Regret$\downarrow$ \\
\midrule
Degradation routing
& 20.7006 & 0.6593 & \textbf{0.3974} & 1.1212 & 48.1 & 3.9618 \\
Direct scorer
& \textbf{22.6932} & 0.6883 & 0.4174 & \textbf{3.4042} & 21.2 & \textbf{1.6788} \\
World model ($H=1$)
& 22.5510 & \textbf{0.6958} & 0.4045 & 3.3364 & \textbf{15.6} & 1.7466 \\
World model ($H=2$)
& 22.4985 & 0.6910 & 0.4094 & 3.2361 & 17.5 & 1.8469 \\
\quad w/o state rollout
& 22.5285 & 0.6938 & 0.4086 & 3.2939 & 16.2 & 1.7891 \\
\quad + harm/severe penalties
& 22.5604 & 0.6909 & 0.4050 & 3.2974 & 18.8 & 1.7857 \\
\bottomrule
\end{tabular}
\end{table}

\subsection{Empirical Risk-Threshold Diagnostic}
\label{sec:supp-risk-calibration}

As a separate diagnostic that no reported controller uses, we estimate
thresholds on the predicted harm probability $p^{\rm harm}$ independently for
each outer fold, using a calibration set $\mathcal D_f$ isolated by
clean-image identity. Candidate thresholds are evaluated by both coverage and
empirical harm rate. For each threshold $\nu_f$, we report the one-sided
Clopper--Pearson upper confidence bound
\begin{equation}
\operatorname{UCB}_{\rm CP}
\left(
\sum_{i\in\mathcal D_f}
\mathbf 1[p^{\rm harm}_i\leq\nu_f]\,y_{{\rm harm},i},
\,
\sum_{i\in\mathcal D_f}
\mathbf 1[p^{\rm harm}_i\leq\nu_f];
\delta
\right).
\label{eq:supp-cp}
\end{equation}
We use an empirical screening target $\epsilon=0.25$ and confidence
parameter $\delta=0.05$. A candidate threshold must cover at least 10
actions and a tool-level threshold at least 20 actions; otherwise, the
procedure falls back to the global threshold. The five calibration folds
contain 229, 215, 216, 215, and 222 transitions (1,097 in total), and the
resulting global thresholds are 0.4953, none (no threshold is eligible),
0.6596, 0.5929, and 0.4563.
No temperature scaling or other recalibration is applied, so a predicted
value such as $p^{\rm harm}=0.2$ should not be read as a $20\%$ harm rate
(see Fig.~\ref{fig:app_sec43}b). These thresholds are empirical references
rather than distribution-free safety guarantees. Because no reported
controller uses them, the negative result for the soft harm penalties in
Table~\ref{tab:policy-ablation} does not bear on threshold-based filtering.

\subsection{Post-Execution Verification}
\label{sec:supp-verifier}

The verifier $V_\psi$ is trained separately from the world model $W_\theta$
and shares no parameters with it or with the degradation estimator $E_\phi$; it
is trained on observed transitions. Given the before/after images $(I_t,\widetilde I_{t+1})$ and the
executed tool, it predicts an acceptance probability $p^{\rm acc}$ and the
Y-PSNR and Y-SSIM gains of the transition; it does not receive the
degradation state or any reference image. In the replays of
Sec.~\ref{sec:closedloop} and Appendix~\ref{app:closedloop}, a candidate is
committed if
\begin{equation}
p^{\rm acc}\geq\eta_f,
\qquad
\widehat{\Delta\mathrm{PSNR}}_Y>0,
\qquad
\widehat{\Delta\mathrm{SSIM}}_Y>-0.005,
\label{eq:supp-commit}
\end{equation}
where $\eta_f$ is a fold-specific threshold selected by $F_{0.5}$ on
inner-validation identities and the hatted gains are the verifier's
predictions, used without an uncertainty margin; otherwise, it is rolled
back to $I_t$. After every commit, the degradation estimator is rerun on the
real output, so subsequent decisions rely on observed rather than imagined
residual degradations. In the online trajectory of Fig.~\ref{fig:trace}, the
rejected action (MAXIM for dehazing) is excluded when the unchanged image is
re-ranked.
The verifier rejects many, though not all, harmful outcomes, and its errors
are partly correlated with the world model's (Appendix~\ref{app:closedloop}).

\FloatBarrier
\section{Additional Candidate-Selection Details}
\label{sec:supp-candidate-value}

\subsection{Scene-Aware Candidate Value Models}

World-model planning first pre-ranks root actions, and only high-value
roots are expanded through real tool execution and second-step
candidate generation. Complementary outputs may additionally be
combined using fixed convex blends.

Each generated candidate is represented using deployment-visible
features, including degradation-state estimates before and after
processing, world-model predictions, tool and trajectory encodings,
self-reference image-change features, frozen visual representations,
and no-reference quality signals such as MANIQA, CLIP-IQA, and MUSIQ.
Reference images are used only to construct training targets and never
enter test-time features.

We train ExtraTrees~\citep{geurts2006extratrees} and a scene-conditioned
MLP using identity-disjoint cross-fitting. The former models
heterogeneous tabular features and non-smooth tool branches, while the
latter captures continuous scene--action interactions. The models
predict the candidate quality vector or relative quality change, which
is converted into target-normalized robust utility for final selection.

For Groups A and B, the final frozen RGB output-level ensemble is
\begin{equation}
I_{\rm ens}
=
0.1875I_{\rm tree}
+
0.5625I_{\rm scene+tree}
+
0.25I_{\rm scene}.
\label{eq:supp-ensemble}
\end{equation}
The weights are frozen from cross-fitted results on the training
identities and remain unchanged at test time.

This post-execution value model serves a different role from the world
model. The candidate-value model operates only after an image has been
generated, whereas the world model predicts action consequences before
execution. Its predicted next state permits counterfactual rollout, but our
ablation (Table~\ref{tab:policy-ablation}) supports only the one-step use of
these predictions.

\FloatBarrier
\section{Additional Qualitative Results}
\label{app:qualitative}

\begin{figure}[h]
\centering
\includegraphics[width=\linewidth]{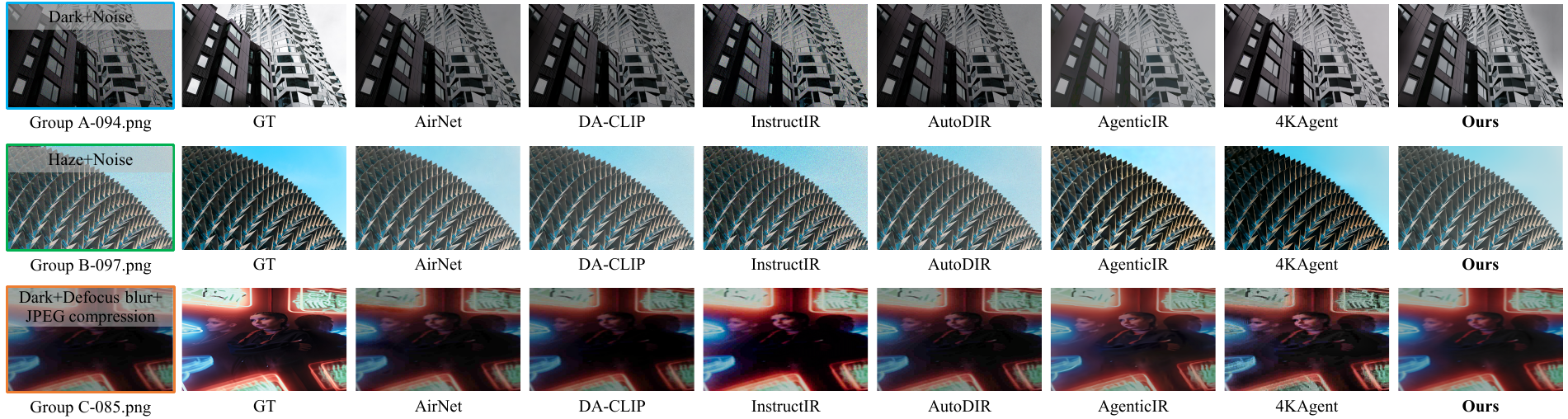}
\caption{Additional qualitative comparisons on MiO100 Groups~A--C.}
\label{fig:supp-qualitative}
\end{figure}

Figure~\ref{fig:supp-qualitative} gives further MiO100 examples from
Groups~A--C in the same format as Fig.~\ref{fig:qualitative_mio100}; the
first column shows the degraded input and its degradation combination.

\FloatBarrier
\section{Open-Loop Ranking: Protocol and Additional Results}
\label{app:sec43}

This appendix complements Sec.~\ref{sec:openloop-ranking}. All results use
the same 160 clean-identity-disjoint OOF states. These states were also used
during policy development, so they serve as a controlled mechanism test rather
than an untouched test-set estimate.

\subsection{Protocol}

\paragraph{Candidate pool.}
For each state, the pool contains every eligible action under the four
degradation targets with the highest OOF presence probability; low resolution is
always among the four targets when the long side of the input is at most 512
pixels. Five slow actions are excluded: MAXIM for haze, motion blur, noise, and
rain, and DiffBIR~\citep{lin2024diffbir} for low resolution. This yields 2,147 executed candidates (13.4 per state
on average). The pool is fixed before any ranker is applied and is not
prescreened by the world model or any scalar scorer. Realized outcomes are
used only for evaluation.

\paragraph{Methods.}
All deployable methods score the same pool from pre-execution inputs and
commit their top-ranked action; ties are broken by action name.
\emph{Tool prior} ranks each tool by its mean realized $\Delta F_Y$ on the 14
fit identities of the fold, ignoring the image and the degradation state.
\emph{Degradation routing} selects the target with the highest predicted
presence and then the tool with the highest mean realized $\Delta F_Y$ for
that target on the fit identities.
The \emph{Direct scorer} is a separately trained network with a single output
head that regresses the same Y-channel fidelity target $\mathbf y_f$ as the
world model's fidelity head (Appendix~\ref{sec:supp-labels}); it shares the
world model's folds, inputs, backbone, training rows, and optimization protocol
(one seed per fold), but is trained with the Gaussian NLL only, without
$\mathcal L_{\rm rank}^{f}$.
The \emph{world model} ranks candidates by the frozen one-step utility
$U_t(a)$ of Eq.~\eqref{eq:action-value}; \emph{WM-F} ranks them by the fidelity
head's prediction $v$ alone. \emph{Oracle} selects the candidate with the highest realized
$\Delta F_Y$. \emph{Random} is the analytic expectation of a uniform choice
within each state.

\paragraph{Outcome labels.}
A selection is \emph{harmful} if $\Delta\mathrm{PSNR}_Y<-0.05$\,dB,
$\Delta\mathrm{SSIM}_Y<-0.005$, or the mean RGB proxy-quality change is
negative, which is exactly the label used to train the harm head
(Appendix~\ref{sec:supp-labels}). It is
\emph{severe} if $\Delta\mathrm{PSNR}_Y\le-1$\,dB or
$\Delta\mathrm{SSIM}_Y\le-0.05$.

\paragraph{Statistics.}
Intervals are 95\% percentile intervals from 10,000 clean-identity
cluster-bootstrap draws that resample the 20 identities with replacement; HR
is recomputed as a ratio of means in every draw. The rank correlation $r_{\rm s}$
is the Spearman correlation between predicted scores and realized
$\Delta F_Y$, computed within each state and then averaged.

\subsection{Full Results}

Table~\ref{tab:sec43_openloop} in the main text reports all top-1 metrics,
including severe harm, LPIPS, within-state rank correlation, and WM-F;
Table~\ref{tab:app_sec43_diff} gives paired differences to the Direct scorer.
Neither world-model score differs significantly from Direct in HR or harm.
Both order candidates within a state better than Direct (the $r_{\rm s}$ intervals
exclude zero), but this does not translate into a better top-1 choice.

\begin{table}[t]
\caption{Paired differences to the Direct scorer with 95\% clean-identity
cluster-bootstrap intervals. Rates are in percentage points.}
\label{tab:app_sec43_diff}
\centering\small
\begin{tabular}{lcc}
\toprule
Metric & World model $-$ Direct & WM-F $-$ Direct \\
\midrule
HR              & $-$2.1 [$-$15.5, 11.8]      & $-$4.8 [$-$15.1, 6.0] \\
Harm            & $-$6.2 [$-$13.8, 1.2]       & $-$2.5 [$-$10.0, 5.0] \\
Severe          & $-$5.6 [$-$16.2, 4.4]       & $-$4.4 [$-$13.8, 5.0] \\
$\Delta$PSNR$_Y$ (dB) & $-$0.13 [$-$0.68, 0.41] & $-$0.21 [$-$0.65, 0.23] \\
$\Delta$SSIM$_Y$ & 0.002 [$-$0.013, 0.019]    & $-$0.002 [$-$0.017, 0.012] \\
$\Delta$LPIPS   & $-$0.013 [$-$0.033, 0.006]  & $-$0.001 [$-$0.018, 0.018] \\
$r_{\rm s}$     & 0.062 [0.009, 0.120]        & 0.067 [0.014, 0.122] \\
\bottomrule
\end{tabular}
\end{table}

Table~\ref{tab:app_sec43_ablation} removes one term of the frozen utility at a
time without re-tuning the others. The fidelity-gain term $v$ drives the
ranking: without it, HR falls to $-14.5\%$. Removing the benefit probability
$p^{\rm ben}$ raises harm from 42.5\% to 46.2\%. Removing $g$, $q$, $d$,
$u$, or $C$ leaves every top-1 metric unchanged; $d$ is identically zero for
the global actions used here (Appendix~\ref{sec:supp-labels}). Removing the
side-effect term $\xi$ slightly improves all top-1 metrics; we nevertheless
report the frozen utility unchanged.

\begin{table}[t]
\caption{Utility-term ablation (one term removed at a time, other coefficients
frozen). Terms follow Eq.~\eqref{eq:action-value}: $v$ fidelity gain,
$p^{\rm ben}$ benefit probability, $\xi$ non-target side effects, $g$
target gain, $q$ proxy-quality gain, $d$ content damage, $u$ uncertainty,
$C$ cost prior.}
\label{tab:app_sec43_ablation}
\centering\small
\setlength{\tabcolsep}{4pt}
\begin{tabular}{lrrrrrrr}
\toprule
Utility & HR (\%) & Harm (\%) & Severe (\%) & $\Delta$PSNR$_Y$
& $\Delta$SSIM$_Y$ & $\Delta$LPIPS & $r_{\rm s}$ \\
\midrule
Full                     & 34.6 & 42.5 & 21.9 &    1.14 &    0.055 & $-$0.069 & 0.309 \\
w/o $v$                  & $-$14.5 & 58.1 & 35.6 & $-$0.66 & $-$0.005 & $-$0.019 & $-$0.092 \\
w/o $p^{\rm ben}$        & 32.4 & 46.2 & 21.9 &    1.08 &    0.050 & $-$0.058 & 0.310 \\
w/o $\xi$                & 36.1 & 41.9 & 20.6 &    1.17 &    0.060 & $-$0.072 & 0.311 \\
w/o $g$, $q$, $d$, $u$, or $C$ & 34.6 & 42.5 & 21.9 & 1.14 & 0.055 & $-$0.069 & 0.309--0.312 \\
\bottomrule
\end{tabular}
\end{table}

\subsection{Ranking Depth, Harm Calibration, and Secondary Scan}

\begin{figure}[t]
\centering
\includegraphics[width=\linewidth]{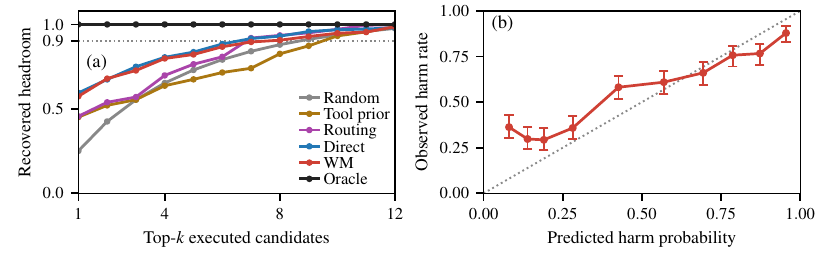}
\caption{Supplementary open-loop diagnostics on the 160 OOF states.
(a) Share of positive oracle headroom recovered by an ideal selector over the
top-$k$ ranked candidates; the dotted line marks 90\%.
(b) Reliability of the world model's harm probability in ten equal-count bins
with Wilson 95\% intervals.}
\label{fig:app_sec43}
\end{figure}

Figure~\ref{fig:app_sec43}a reports the share of positive oracle headroom
obtained when an ideal selector keeps the best of the top-$k$ ranked
candidates. The smallest $k$ reaching 90\% is 7 for routing and Direct, 8 for
the world model, 9 for Random, and 10 for Tool prior.
Figure~\ref{fig:app_sec43}b shows the reliability of the world model's harm
probability over ten equal-count bins (ECE 0.106, Brier 0.222). No
recalibration is applied.
In the lowest-probability bins the observed harm rate (about 0.3) exceeds the
prediction, so the probabilities should be read as a ranking signal rather
than as calibrated risks.

Paired AUC differences for the world-model harm probability are $+0.118$
$[0.091, 0.146]$ versus negative Direct gain and $+0.073$ $[0.054, 0.093]$
versus negative WM-F gain. In Fig.~\ref{fig:sec43_openloop}b, intervals are
evaluated on a 0.5\% grid of harm rates. At matched harm rates of 5\%, 10\%,
and 15\%, WM-$\tau$ minus Direct-$\tau$ HR is $+0.0$ $[-11.0, 11.9]$, $+3.5$
$[-10.4, 19.3]$, and $+10.7$ $[-7.8, 21.2]$ points.

The pre-declared secondary scan WM-$\lambda$ subtracts
$\lambda_{\rm harm}p^{\rm harm}$ from the utility and abstains when the maximum
utility is negative. It attains higher HR than both $\tau$-scans at harm rates
above about 25\%, but its range does not reach the matched rates of 5--15\%,
so it is not used for inference.

\FloatBarrier
\section{Closed-Loop Replay: Additional Results}
\label{app:closedloop}

\paragraph{Paired differences.}
Paired \method-minus-reactive differences (95\% clean-identity
cluster-bootstrap intervals) are
$-1.26$ calls/image $[-1.50,-1.02]$, $-1.49$ rejected calls/image
$[-1.78,-1.21]$, $-14.4$\,s tool time $[-24.7,-4.2]$, $+0.92$\,dB Y-PSNR
$[0.49,1.34]$, $+0.015$ Y-SSIM $[0.003,0.027]$, and $-23.1$ points no-commit
rate $[-30.6,-15.6]$; LPIPS and harm differences include zero. Against
fidelity-head ranking, \method reduces harm by 3.75 points (95\% CI for the
reduction: 0.6--7.5), while the PSNR, SSIM, and LPIPS differences include zero.

\begin{figure}[t]
\centering
\includegraphics[width=0.9\linewidth]{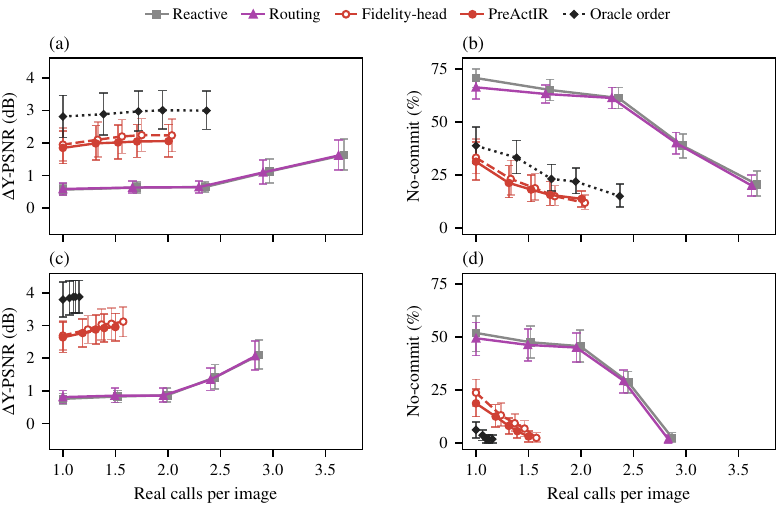}
\caption{Quality versus real tool calls on the 160 OOF states with a shared
candidate pool. Top: learned verifier; bottom: ground-truth verifier that
commits exactly the candidates labeled beneficial
(Appendix~\ref{sec:supp-labels}). Points from left to right correspond to
$B=1,2,3,4,6$; axes are shared across rows. (a,c) Y-PSNR gain over the input.
(b,d) Share of states with no committed candidate within budget. Error bars
are 95\% clean-identity cluster-bootstrap intervals; the oracle order uses
realized outcomes and is shown only as a reference.}
\label{fig:app_closedloop}
\end{figure}

\paragraph{Budget.}
Table~\ref{tab:quality_cost} reports \method for $B\in\{1,2,4\}$, and
Figure~\ref{fig:app_closedloop} (top) compares all policies for
$B\in\{1,2,3,4,6\}$, with the pool and verifier fixed. At every budget,
pre-execution ranking reaches a given Y-PSNR gain with fewer real calls than
reactive trial-and-error and routing, largely because it finds a verified candidate
earlier (panel b). For \method, harm rises from 8.8\% to 14.4\% between
$B=1$ and $B=4$ as the no-commit rate falls from 31.2\% to 15.6\%, so the
lower harm at small budgets largely reflects abstention.

\begin{table}[t]
\caption{Sensitivity of \method to the real-execution budget on the 160 OOF
states, with the candidate pool and verifier fixed. Harm counts no-commit
outcomes as non-harmful; time counts restoration-tool execution only. The
$B=4$ row is the \method row of Table~\ref{tab:matched-reactive}.}
\label{tab:quality_cost}
\centering\small
\setlength{\tabcolsep}{4pt}
\begin{tabular}{lccccccc}
\toprule
Budget & PSNR$_Y\uparrow$ & SSIM$_Y\uparrow$ & LPIPS$\downarrow$
& Harm (\%)$\downarrow$ & No commit (\%)$\downarrow$ & Calls$\downarrow$
& Time (s)$\downarrow$ \\
\midrule
$B=1$ & 21.88 & 0.683 & 0.423 &  8.8 & 31.2 & 1.00 & 18.5 \\
$B=2$ & 22.02 & 0.687 & 0.419 & 11.2 & 21.2 & 1.31 & 21.0 \\
$B=4$ & 22.08 & 0.690 & 0.416 & 14.4 & 15.6 & 1.71 & 25.3 \\
\bottomrule
\end{tabular}
\end{table}

\paragraph{Routing and the two candidate pools.}
Routing fits its tool order on the 14 fit identities of each fold, rather than
on the two inner identities used by reactive trial-and-error, yet it performs
almost identically to it. The contrast with
Sec.~\ref{sec:openloop-ranking} reflects two differences between the
settings: the open-loop test uses estimated beliefs and a belief-based pool,
whereas the closed-loop replay uses annotated states and an annotation-based
pool that includes the five slow actions excluded from the open-loop pool.
Routing's first choice coincides with its open-loop choice in only 34 of 160
states. Because the world model is trained on annotated states, estimated
beliefs introduce an input shift, and the larger advantage of pre-execution
ranking in the closed loop may partly reflect the accurate state.

\paragraph{Verifier agreement.}
At $B=1$, the learned verifier accepts 72\% of the beneficial first choices
of \method but only 57\% of those of routing, and it also accepts 62\% of
\method's harmful first choices versus 11\% of routing's
(Table~\ref{tab:app_verifier}). Its errors are therefore correlated with the
world model's ranking. To test whether the closed-loop advantage merely
reflects this agreement, we replay all policies with a ground-truth verifier
(Fig.~\ref{fig:app_closedloop}, bottom). The gap widens rather than closes:
at $B=4$, routing needs 1.01 more calls per image (CI $[0.75, 1.26]$) and
gains 1.58\,dB less (CI $[-1.89,-1.28]$) than \method, compared with 1.20
calls and 0.95\,dB under the learned verifier. The ordering advantage is thus
not an artifact of verifier agreement, but the learned verifier catches fewer
of the world model's harmful choices than of routing's.

\begin{table}[t]
\caption{Learned-verifier decisions on each policy's first choice at $B=1$,
using the cached ground-truth labels.}
\label{tab:app_verifier}
\centering\small
\begin{tabular}{lrrr}
\toprule
Policy & Accepted & Beneficial accepted & Harmful accepted \\
\midrule
Reactive       &  47/160 & 43/77 (55.8\%)  &  4/78 (5.1\%) \\
Routing        &  54/160 & 46/81 (56.8\%)  &  8/74 (10.8\%) \\
Fidelity-head  & 107/160 & 88/122 (72.1\%) & 19/34 (55.9\%) \\
\method        & 110/160 & 94/130 (72.3\%) & 16/26 (61.5\%) \\
\textit{Oracle order} & 98/160 & 97/150 (64.7\%) & 1/7 (14.3\%) \\
\bottomrule
\end{tabular}
\end{table}

\FloatBarrier
\section{Candidate-Frontier Audit and Targeted Tool Adaptation}
\label{sec:supp-tool-evolution}

\subsection{Candidate-Frontier Audit}
\label{sec:supp-frontier}

Sec.~\ref{sec:frontier} solves Eq.~\eqref{eq:frontier-audit} exactly as a
mixed-integer program that selects one frozen option per state and
constrains dataset averages. The audit is post hoc and reference-assisted:
it reads test references only to score frozen options, never changes a
candidate or model, and is not a deployable policy.

A descriptive distortion--perception trend fitted across the outputs of
different controllers can motivate such an audit, but it cannot establish
that the candidate space is insufficient; only the exact program can. For
the target check below, the options comprise the 21,385 protocol-compatible
candidates of the complete 33-action Group~C pool and the outputs of four
fixed-pool controller configurations (F1--F4; 1,600 outputs), i.e., 22,985
options for 400 states. The pre-specified Group~C targets (PSNR $20.59$,
SSIM $0.6039$, MANIQA $0.3047$, CLIP-IQA $0.4503$, and MUSIQ $54.57$) are
jointly attainable. After raising them by
$0.01/0.0001/0.001/0.001/0.01$, respectively, the PSNR-maximizing selection
subject to the three perceptual floors attains
$21.79/0.617/0.325/0.460/54.58$, whereas interpolating the four controller
aggregates F1--F4 at the target MUSIQ level gives only $20.05/0.571$
Y-PSNR/Y-SSIM. The fixed pool is therefore not the binding constraint.

\subsection{External Replay Construction}
\label{sec:supp-replay}

To expand the Group C candidate frontier, we construct 4,000 training
pairs from external images using the official AgenticIR degradation
operators. The replay set consists of

\begin{itemize}
    \item 2,518 rain+noise+$\times4$ degradation pairs;
    \item 1,228 clean-image $\times4$ pairs; and
    \item 254 real paired hazy images with additional official
    motion-blur and $\times4$ degradation.
\end{itemize}

No MiO100 HQ image is used during tool adaptation.

\subsection{Replay-Adapted Super-Resolution Tool}
\label{sec:supp-hat}

We use HAT-S~\citep{chen2023hat} as the expandable
super-resolution tool and further train it for 4,000 iterations on the
external replay data.

On a fixed external validation set of 120 pairs, Y-PSNR/Y-SSIM improve from
24.84/0.7432 before adaptation to 26.85/0.7692 after 4,000 iterations.

The frozen adapted tool is applied only to the 200 Group C states that
contain low resolution; other states retain the original
high-perceptual-quality candidates.

\subsection{Final Group C Candidate Fusion}
\label{sec:supp-groupc-fusion}

For Group C, the final candidate is constructed as
\begin{equation}
I_C
=
(1-\alpha)I_p
+\alpha I_{\rm sr}
+\kappa
\left(
I_p-G_\sigma(I_p)
\right),
\label{eq:supp-groupc-fusion}
\end{equation}
where $I_p$ denotes the original high-perceptual-quality candidate,
$I_{\rm sr}$ the replay-adapted super-resolution candidate, and
$G_\sigma$ a Gaussian smoothing operator.

We use
\[
\alpha=0.62,
\qquad
\kappa=0.12,
\qquad
\sigma=1.5.
\]
This expanded candidate moves the deployed Group~C operating point beyond
the descriptive F1--F4 trend (Appendix~\ref{sec:supp-frontier}). It does not
exceed the reference-assisted fixed-pool oracle, so it demonstrates a
practical intervention rather than a necessary capacity expansion.

\FloatBarrier

\end{document}